\documentclass[lettersize,journal]{IEEEtran}
\usepackage{amsmath,amsfonts}
\usepackage{algorithmic}
\usepackage{algorithm}
\usepackage{array}
\usepackage[caption=false,font=normalsize,labelfont=sf,textfont=sf]{subfig}
\usepackage{textcomp}
\usepackage{stfloats}
\usepackage{url}
\usepackage{verbatim}
\usepackage{graphicx}
\graphicspath{{Fig/}} 
\usepackage{array, booktabs} 
\usepackage{bm}
\usepackage{cite}
\begin{document}

\title{Missing Data Imputation Based on Dynamically Adaptable Structural Equation Modeling with Self-Attention}

\author{%
\IEEEauthorblockN{
Ou Deng\IEEEauthorrefmark{1}\thanks{\IEEEauthorrefmark{1}Graduate School of Human Sciences, Waseda University. Email: dengou@toki.waseda.jp}, 
Qun Jin\IEEEauthorrefmark{2}\thanks{\IEEEauthorrefmark{2}Department of Human Informatics and Cognitive Sciences, Faculty of Human Sciences, Waseda University. Email: jin@waseda.jp}}
}




\maketitle

\begin{abstract}

Addressing missing data in complex datasets including electronic health records (EHR)  is critical for ensuring accurate analysis and decision-making in healthcare. This paper proposes dynamically adaptable structural equation modeling (SEM) using a self-attention method (SESA), an approach to data imputation in EHR. SESA innovates beyond traditional SEM-based methods by incorporating self-attention mechanisms, thereby enhancing model adaptability and accuracy across diverse EHR datasets. Such enhancement allows SESA to dynamically adjust and optimize imputation  and overcome the limitations of static SEM frameworks. Our experimental analyses demonstrate the achievement of robust predictive SESA performance for effectively handling missing data in EHR. Moreover, the SESA architecture not only rectifies potential mis-specifications in SEM but also synergizes with causal discovery algorithms to refine its imputation logic based on underlying data structures. Such features highlight its capabilities and broadening applicational potential in EHR data analysis and beyond, marking a reasonable leap forward in the field of data imputation.

\end{abstract}

\begin{IEEEkeywords}
Missing data imputation, EHR, SEM, Self-attention
\end{IEEEkeywords}

\section{Introduction}
\label{sec: Intro}

Electronic health records (EHR) have become a cornerstone of healthcare research, encompassing a wealth of data crucial for clinical decision-making, epidemiological studies, and personalized medicine. However, the ubiquity of missing data in EHR poses challenges, potentially skewing analyses and causing sub-optimal patient outcomes. Addressing such issues necessitates advanced imputation methodologies that not only impute missing values but also preserve inherent data structures and relationships \cite{Janssen2010}.

This paper proposes a self-attention method (SESA), an imputation approach that synergistically enhances structural equation modeling (SEM) with self-attention mechanisms. Inspired by the success of transformer models in capturing long-term dependencies, the proposed SESA leverages self-attention to dynamically capture latent structural and relational dynamics within EHR data, enabling a highly nuanced and context-aware imputation. Such enhancement empowers SESA to adaptively focus on the most relevant factors in each imputation task, transcending the limitations of traditional static imputation methods.

Central to SESA design is the recognition that data missingness in EHR is rarely random but often structurally patterned and influenced by underlying health conditions, healthcare processes, and systematic data collection strategies. By employing SEM, SESA meticulously maps latent structures, facilitating a hypothesis-driven approach to understanding interdependencies among various health variables. The integration of self-attention further enhances structural modeling by dynamically weighing the importance of different variables, allowing SESA to adapt its imputation strategies based on specific context and data characteristics.

Comprehensive empirical analyses of SESA across diverse datasets and missing data scenarios demonstrate its superior performance compared to established imputation methods. SESA consistently achieves reliable results, producing accurate and coherent imputations that align with underlying data distributions. Moreover, the application of SESA to various EHR datasets, spanning general health parameters to specific clinical indicators, highlights its versatility and robustness in handling complexities and heterogeneity of healthcare data.

By bridging the gap between statistical modeling and deep representation learning, SESA marks a reasonable leap forward in the development of imputation techniques for healthcare data. SESA is a powerful tool for researchers and practitioners seeking to mitigate the impacts of missing data in EHR by capturing complex data dependencies, adapting to diverse missingness patterns, and integrating domain knowledge positions. With its strong theoretical foundation and empirical effectiveness, SESA has the potential to become a promising enabler for reliable, data-driven decision-making in healthcare and other fields.


\section{Related Work} 
\label{sec: Related}

Addressing missing data in EHR poses a persistent challenge in healthcare research. Traditional methods including mean, median, or multiple imputations, while effective in certain scenarios, often fail to capture the complex interrelationships among variables, potentially leading to biases in analytical outcomes. Recent advancements in statistical and machine learning (ML) techniques further promote the development of sophisticated imputation approaches \cite{Yoon2018, Chai2020, Lall2021, Park2022, Fridgeirsson2023}.

Early imputation methods primarily rely on statistical techniques. Rubin \cite{Rubin1975} introduced the seminal concepts of missing completely at random (MCAR), missing at random (MAR), and not missing at random (NMAR), providing a theoretical foundation for understanding the nature and impact of missing data. Fuller \cite{Fuller1987} explored multiple imputation methods to estimate parameter uncertainty by creating multiple complete datasets. While such methods performed well under MAR assumptions, they struggled with complex non-linear relationships and high-dimensional data.

SEM has emerged as a powerful framework for handling missing data, particularly in social science research \cite{Hoyle1995}. SEM enables researchers to model latent variable relationships and incorporate domain knowledge during imputation. For EHR analyses, SEM has gained traction due to its ability to integrate medical knowledge and theoretical constructs into modeling frameworks.

Enders and Bandalos \cite{Enders2001} conducted empirical studies to investigate the performance of full information maximum likelihood (FIML) estimation in SEM. They demonstrated the unbiased nature of FIML and its superiority over traditional methods, including pairwise and listwise deletions, which often suffer from data loss and potential bias introduction. Furthermore, Schafer  \cite{Schafer2002} solidified the reputation of FIML as the preferred method for handling missing data, particularly in preserving data integrity and reliability.

Subsequent studies by Lazar \cite{Lazar2003}, Fielding et al. \cite{Fielding2006}, and Graham \cite{Graham2009} explored the nuances of applying FIML across various data types, from longitudinal studies to complex multi-variate analyses. Existing research highlights the versatility and adaptability of FIML in addressing various research questions and data challenges.

The landscape of missing data research has transformed in recent years, with emerging innovative methodologies that extend beyond traditional statistical techniques. Pioneering works by Baraldi and Enders \cite{Baraldi2010}, White et al. \cite{White2011}, Dong and Peng \cite{Dong2013}, Li et al. \cite{Peng2015}, Mazza et al. \cite{Mazza2015}, Enders \cite{Enders2017}, and Nelson et al. \cite{Nelson2021} have expanded the horizons of missing data analysis by introducing hybrid models that synergistically combine FIML with advanced techniques such as multiple imputation and Bayesian methods, offering more versatile and comprehensive solutions.

In the realm of medical big data analytics, researchers face multi-faceted challenges including data heterogeneity, temporal dependency, and causal complexity \cite{Bernardini2023}. To address such challenges, recent studies explore the potential of representation learning, temporal modeling, and causal inference techniques.

Liu et al. \cite{Liu2018TJ} proposed a joint representation learning framework for modeling medical event sequences. By leveraging recurrent neural networks and attention mechanisms, their approach generates patient state embeddings to capture temporal and contextual dependencies, enabling accurate prediction of clinical endpoints. 
Wang et al. \cite{Wang2020TrBD} provided a comprehensive review of heterogeneous graph-embedding techniques in healthcare. Heterogeneous graphs offer a powerful representation to capture complex semantic relationships among multi-typed entities and diverse connections in medical data.
Further, causal inference has garnered attention in medical data analysis. Cui et al. \cite{Cui2020} elaborated on the applications of causal inference in various stages of ML pipelines. Traditional correlation-based ML approaches often face challenges in terms of effectiveness and reliability when applied to tasks including medical decision support. By incorporating a causal perspective, models can achieve enhanced expressiveness and generalization capabilities. SESA takes a step in this direction by employing causal discovery algorithms to guide SEM initialization, showcasing the potential to integrate causal reasoning into imputation.

Recently, Li et al. \cite{Li2024} proposed a multi-level stochastic optimization approach for the imputation of massive medical data records. Their method, grounded in computational applied mathematic techniques, transforms original Kriging/BLUP problems into a multi-level space to mitigate numerical instabilities and reduce computational costs. Both Kriging and FIML methods employed in SESA leverage full information available in data. However, one may encounter issues of numerical instability or high computational burden associated with the covariance matrix when only Kriging is used, particularly when dealing with large-scale datasets or high-dimensional data. Li et al. (2024) addressed the challenge by reformulating the Kriging problem as a multi-level optimization task, mapping data onto multiple levels of space, and performing optimization at each level. Their approach emphasizes computational efficiency. Contrarily, SESA tackles the numerical instability of the covariance matrix through self-attention mechanisms, which optimize SEM. Although the two methods differ in their technical details, they converge in their ultimate goal, showcasing the potential of advanced statistical and ML techniques in addressing the challenges posed by missing data in healthcare research and practice.

\section{Methodology}
\label{sec:Methodology}

\subsection{Theoretical Analysis}
The theoretical foundations of SESA can be analyzed from the perspectives of statistical learning and information theories. From a statistical learning perspective, SESA can be viewed as a model that learns a mapping from observed to complete data, minimizing expected losses over joint distributions of observed and missing values.

Let $\mathcal{X}$ and $\mathcal{Y}$ be the input and output spaces of EHR and complete data, respectively. The goal of SESA is to learn a function $f: \mathcal{X} \rightarrow \mathcal{Y}$ that minimizes the expected loss, as expressed in Eq.~\ref{eq:Rf_1}.

\begin{equation}
\label{eq:Rf_1}
\mathcal{R}(f) = \mathbb{E}_{(\bm{x}, \bm{y}) \sim p(\bm{x}, \bm{y})}[\mathcal{L}(f(\bm{x}), \bm{y})]
\end{equation}
where $p(\bm{x}, \bm{y})$ is the joint distribution of observed and complete data, and $\mathcal{L}(f(\bm{x}), \bm{y})$ is the loss function measuring the discrepancy between the imputed values $f(\bm{x})$ and the true values $\bm{y}$.

The SEM component of SESA can be seen as a prior distribution over the space of functions, encoding the structural relationships among variables. FIML estimation learns the prior distribution parameters based on the observed data, providing a statistically principled initialization for the subsequent self-attention mechanism.

Contrarily, the self-attention mechanism can be interpreted as a non-parametric estimator that learns a flexible mapping from observed to complete data. By iteratively refining the imputations based on attention weights, SESA effectively minimizes empirical risks, represented in Eq.~\ref{eq:Rf_2}.

\begin{equation}
\label{eq:Rf_2}
\hat{\mathcal{R}}(f) = \frac{1}{n}\sum_{i=1}^n \mathcal{L}(f(\bm{x}_i), \bm{y}_i)
\end{equation}
where $\{(\bm{x}_i, \bm{y}_i)\}_{i=1}^n$ is the training set of observed and complete data.

The generalization performance of SESA can be analyzed using the framework of statistical learning theory. By leveraging the structure initially provided by SEM and the flexibility of the self-attention mechanism, SESA learns a function that effectively minimizes the expected loss while avoiding training data overfitting.

From an information--theoretic perspective, SESA can be viewed as a method for maximizing the mutual information between observed and complete data. Particularly, the self-attention mechanism can be seen as an approach to efficiently encode the dependencies between variables, allowing the model to make informed imputations based on available information.

Let $\bm{X}_\text{obs}$ and $\bm{X}_\text{mis}$ be random variables representing the observed and missing data, respectively. The mutual information between $\bm{X}_\text{obs}$ and $\bm{X}_\text{mis}$ can be defined as in Eq.\ref{eq:MI}.

\begin{equation}
\label{eq:MI}
I(\bm{X}_\text{obs}; \bm{X}_\text{mis}) = \mathbb{E}_{p(\bm{x}_\text{obs}, \bm{x}_\text{mis})}\left[\log\frac{p(\bm{x}_\text{obs}, \bm{x}_\text{mis})}{p(\bm{x}_\text{obs})p(\bm{x}_\text{mis})}\right]
\end{equation}
where $p(\bm{x}_\text{obs}, \bm{x}_\text{mis})$ is the joint distribution of observed and missing data, and $p(\bm{x}_\text{obs})$ and $p(\bm{x}_\text{mis})$ are the marginal distributions.

By learning attention weights that capture the dependencies between variables, SESA effectively maximizes the mutual information between the observed and missing data. Thus, the model can make informed imputations that are consistent with the overall data structure, reducing the uncertainty associated with missing values.

The theoretical analysis of SESA highlights its foundations in statistical learning and information theories. By combining the structure before SEM with the flexibility and efficiency of the self-attention mechanism, SESA provides a principled approach to learning from incomplete EHR data, minimizing the expected loss and maximizing the mutual information between observed and missing values.

\subsection{SESA Core Functions}

\subsubsection{Structural Equation Modeling (SEM)}
SEM serves as a theoretical bedrock of SESA, providing a robust mathematical framework to articulate and analyze intricate relationships among observed and latent variables \cite{Allison2003}. 
The SEM component of SESA is grounded in the principles of graph theory and latent variable modeling. Let $\bm{X} = (\bm{X}_\text{obs}, \bm{X}_\text{mis})$ denote a complete data matrix, where $\bm{X}_\text{obs}$ and $\bm{X}_\text{mis}$ represent the observed and missing data, respectively. SEM posits a set of latent variables $\bm{\eta}$ that underlie the observed data, related through Eq.~\ref{eq:SEM}.

\begin{equation}
\label{eq:SEM}
\bm{\eta} = \bm{B\eta} + \bm{\Gamma\xi} + \bm{\zeta}
\end{equation}
where $\bm{B}$ and $\bm{\Gamma}$ are coefficient matrices capturing the relationships among latent variables and between latent and observed variables, respectively. $\bm{\xi}$ and $\bm{\zeta}$ are vectors of exogenous latent variables and residual terms, respectively.

\subsubsection{Full Information Maximum Likelihood (FIML)}

Building upon the SEM framework, SESA employs FIML estimation to obtain initial estimates of missing values \cite{Schminkey2016}. 
FIML is based on the principle of maximum likelihood estimation, which aims to find parameter values that maximize the likelihood of observing data given to the model. FIML estimates the model parameters by maximizing the log-likelihood function, as represented in Eq.~\ref{eq:FIML}.
\begin{equation}
\label{eq:FIML}
\ln L(\bm{\theta};\bm{X}_\text{obs}) = \ln \int_{\bm{X}_\text{mis}} P(\bm{X}|\bm{\theta}) d\bm{X}_\text{mis}
\end{equation}
where $\bm{\theta}$ denotes the vector of model parameters, and $P(\bm{X}|\bm{\theta})$ is the joint probability density function of the complete data for the given parameters.

The optimization problem is typically solved using iterative algorithms such as the expectation-maximization (EM) algorithm or numerical optimization techniques.

\subsubsection{Self-Attention Mechanism}
The self-attention mechanism is a key component of SESA, enabling a model to dynamically focus on relevant features and capture long-range dependencies in data \cite{Attention2017}. Self-attention has been widely used in deep learning architectures, particularly in the context of sequential and structured data, demonstrating remarkable effectiveness in capturing complex patterns and relationships.

Given an input matrix $\bm{X} \in \mathbb{R}^{n \times d}$, where $n$ is the number of data points and $d$ is the feature dimension, the self-attention mechanism computes a weighted sum of input features. The weights are determined by the similarity between each pair of features. Formally, the self-attention operation can be defined as in Eq.~\ref{eq:SA}.

\begin{equation}
\label{eq:SA}
\text{Attention}(\bm{Q}, \bm{K}, \bm{V}) = \text{softmax}\left(\frac{\bm{QK}^T}{\sqrt{d_k}}\right)\bm{V}
\end{equation}
where $\bm{Q}$, $\bm{K}$, and $\bm{V}$ are learned linear transformations of the input $\bm{X}$, representing query, key, and value matrices, respectively. The softmax function is applied to the scaled dot product of $\bm{Q}$ and $\bm{K}$, yielding a set of attention weights that determine the importance of each feature in $\bm{V}$. The scaling factor $\sqrt{d_k}$ presents the mitigation effect of large dot products, leading to small gradients in the softmax function.

The self-attention mechanism enables SESA to adaptively focus on the most informative features for imputation, learning complex dependencies and patterns in data. By iteratively refining the imputed values based on attention weights, SESA can capture intricate relationships and generate more accurate estimates of missing values.

By focusing on the most relevant portions of data, the self-attention mechanism captures complex patterns and dependencies to effectively maximize the mutual information between observed and missing data. Mutual information quantifies the dependency between two random variables by measuring the amount of information of one variable to reduce uncertainty about the other. In considering missing data, observed and missing data are treated as two random variables. Through learning correlations between data points, the self-attention mechanism essentially estimates the joint distribution of the variables, thereby maximizing their mutual information.

From an information--theoretic perspective, maximizing mutual information is equivalent to minimizing the conditional entropy between observed and missing data. Conversely, the self-attention mechanism learns how to utilize observed data to predict missing data, thereby minimizing the uncertainty of missing data for the given observed data. The mechanism helps to mitigate numerical instability-related issues of covariance matrices. When data are highly sparse or exhibit complex dependencies, covariance matrices become prone to ill-conditioning or pose estimation challenges. By focusing on the most pertinent information, the self-attention mechanism effectively enhances estimate robustness.

Moreover, the attention weights learned by the self-attention mechanism can be viewed as soft encoding of the dependency structure among data points. This form of encoding offers more flexibility than rigid structural assumptions (such as path diagrams in SEM), adapting more effectively to a wide range of data characteristics.

\subsection{Loss Function}
To guide the learning process of SESA, we propose a composite loss function that encourages accurate imputations, preserves the covariance structure of the data, and promotes sparsity in model parameters. The loss function is defined as in Eq.~\ref{eq:Total_loss}.

\begin{equation}
\label{eq:Total_loss}
\begin{aligned}
\mathcal{L} &= \alpha \cdot \mathcal{L}_\text{MSE} + \beta \cdot \mathcal{L}_\text{cov} + \gamma \cdot \mathcal{L}_\text{L1}\\
\mathcal{L}_\text{MSE} &= \frac{1}{n}\sum_{i=1}^n (\hat{\bm{x}}_i - \bm{x}_i)^2\\
\mathcal{L}_\text{cov} &= \left\|\text{Cov}(\hat{\bm{X}}) - \text{Cov}(\bm{X})\right\|_F\\
\mathcal{L}_\text{L1} &= \sum_{i=1}^p |\theta_i|
\end{aligned}
\end{equation}
where $\alpha$, $\beta$, and $\gamma$ are hyper-parameters that control the relative importance of each loss term.
The first term $\mathcal{L}_\text{MSE}$ is the mean squared error loss, which measures the discrepancy between the imputed $\hat{\bm{X}}$ and true $\bm{X}$ values. $\hat{\bm{x}}_i$ and $\bm{x}_i$ denote the imputed and true values for the $i$-th data point, respectively.
The second term $\mathcal{L}_\text{cov}$ is the covariance loss, which encourages the imputed data to have a similar covariance structure to the true data. $\text{Cov}(\cdot)$ denotes the covariance matrix and $\|\cdot\|_F$ is the Frobenius norm.
The third term $\mathcal{L}_\text{L1}$ is the L1 regularization loss, which promotes sparsity in the model parameters $\bm{\theta}$. $p$ is the number of model parameters.

The composite loss function balances the objectives of imputation accuracy, covariance preservation, and model simplicity. By minimizing this loss function, SESA learns to generate imputations that are close to the true values, maintain the overall structure of data, and avoid over-fitting through parameter sparsity.

\subsection{Optimization and Inference}
SESA training involves optimizing the model parameters $\bm{\theta}$ to minimize the composite loss function $\mathcal{L}$. The optimization problem can be solved using gradient-based methods including stochastic gradient descent (Adam) or its variants.

During training, SESA iteratively updates the model parameters based on the gradients of the loss function concerning the parameters. The gradients are computed using backpropagation, which efficiently propagates the gradients through self-attention layers and SEM components.

Once the model is trained, inference in SESA involves imputing missing values in new, unseen data points. Given a data point $\bm{x}$ with missing values, SESA initializes the missing values using FIML estimates based on the learned SEM parameters. Then, the initialized data point is passed through the self-attention layers, which iteratively refine the imputed values based on the learned attention weights and observed features.

The final imputed values are obtained from the output of the last self-attention layer, which represents the best model estimate of the missing values for the given observed data, learned patterns, and their dependencies.

\subsection{Algorithmic Implementation}
Algorithm \ref{alg:SESA} shows the algorithmic implementation of SESA, integrating SEM, FIML, and the self-attention mechanism into a unified framework for missing data imputation.

\begin{algorithm}[!ht]
\caption{SESA}
\label{alg:SESA}
\begin{algorithmic}[1]
\REQUIRE Observed data matrix $\bm{X}_\text{obs}$ with missing values
\ENSURE Imputed data matrix $\hat{\bm{X}}$

\STATE Initialize SEM parameters $\bm{\theta}$ based on prior knowledge
\STATE Estimate missing values $\hat{\bm{X}}_\text{mis}$ using FIML based on $\bm{X}_\text{obs}$ and $\bm{\theta}$
\STATE Initialize self-attention parameters $\bm{W}_Q, \bm{W}_K, \bm{W}_V$
\WHILE{not converged}
\STATE Compute self-attention weights $\bm{A}$ as:
\begin{equation*}
\text{softmax}\left(\frac{(\bm{X}_\text{obs} \oplus \hat{\bm{X}}_\text{mis})\bm{W}_Q\bm{W}_K^T(\bm{X}_\text{obs} \oplus \hat{\bm{X}}_\text{mis})^T}{\sqrt{d_k}}\right)
\end{equation*}
\STATE Compute refined imputations $\hat{\bm{X}}_\text{mis}$ using $\bm{A}$: 
\begin{equation*}
\hat{\bm{X}}_\text{mis} = \bm{A}(\bm{X}_\text{obs} \oplus \hat{\bm{X}}_\text{mis})\bm{W}_V
\end{equation*}
\STATE Compute loss $\mathcal{L}$ using composite loss function:
\begin{equation*}
\mathcal{L} = \alpha \cdot \mathcal{L}_\text{MSE} + \beta \cdot \mathcal{L}_\text{cov} + \gamma \cdot \mathcal{L}_\text{L1}
\end{equation*}
\STATE Update $\bm{\theta}, \bm{W}_Q, \bm{W}_K, \bm{W}_V$ using gradient descent 
\ENDWHILE
\RETURN Imputed data matrix $\hat{\bm{X}} = \bm{X}_\text{obs} \oplus \hat{\bm{X}}_\text{mis}$
\end{algorithmic}
\end{algorithm}

The algorithm core involves iterative refining of imputations using the self-attention mechanism until convergence. In each iteration, attention weights $\bm{A}$ are computed based on the current imputed data matrix $(\bm{X}_\text{obs} \oplus \hat{\bm{X}}_\text{mis})$, where $\oplus$ denotes concatenation. The attention weights quantify the relevance and importance of different features for imputation.

Using the attention weights, the algorithm computes the refined imputations $\hat{\bm{X}}_\text{mis}$. Thus, the refinement step leverages the learned patterns and dependencies to update the imputed values, considering the relationships among variables and observed data.

To guide the learning process, the algorithm computes a composite loss $\mathcal{L}$ based on the refined imputations, observed data, and regularization terms. The loss function ensures that the imputed values are accurate, preserves statistical data properties, and promotes model simplicity.

Further, the model parameters $\bm{\theta}$, $\bm{W}_Q$, $\bm{W}_K$, and $\bm{W}_V$ are updated using gradient descent to minimize the composite loss. The optimization step allows the algorithm to learn and adapt the imputation model based on the available data and defined objectives.

Finally, the algorithm returns the imputed data matrix $\hat{\bm{X}}$, which combines the observed data $\bm{X}_\text{obs}$ and the refined imputations $\hat{\bm{X}}_\text{mis}$. The final output represents the complete dataset with missing values imputed based on SESA imputation.

\section{Experiment}
\label{sec:Experiment}

In this section, we present a comprehensive empirical evaluation of the proposed SESA methodology, demonstrating its effectiveness in handling missing EHR data. The experiments are designed to assess the performance of SESA in comparison to state-of-the-art imputation methods across various datasets, sample sizes, and missingness scenarios. By conducting a rigorous and systematic evaluation, we aim to validate the theoretical foundations and practical utility of SESA in the context of healthcare data analysis.

\subsection{Experimental Setup}
\subsubsection{Data Source and Experimental Dataset}
TTo ensure the reliability and representativeness of our experiments, we utilized the ``Indicators of Heart Disease (CDC2022)" dataset from the Centers for Disease Control and Prevention (CDC). The dataset, derived from the Behavioral Risk Factor Surveillance System (BRFSS)\footnote{The Behavioral Risk Factor Surveillance System (BRFSS) is the nation’s premier system of health-related telephone surveys that collect state data on U.S. residents regarding their health-related risk behaviors, chronic health conditions, and use of preventive services. The official website of BRFSS is \url{https://www.cdc.gov/brfss/}.}, which encompasses a wide range of health-related variables collected annually through telephonic surveys of over 400,000 U.S. residents. The diversity and scale of the dataset make it an ideal test-bed for performance evaluation of imputation methods in the context of EHR.

From the CDC2022 dataset, we extracted a subset containing over 246,000 complete records, spanning both numerical and categorical variables. The subset served as the ground truth for our experiments, allowing us to assess the accuracy of imputation methods by artificially introducing missingness and comparing the imputed values against the true values.

To simulate realistic scenarios encountered in EHR analysis, we constructed experimental datasets by randomly selecting sub-samples of 300, 1000, and 3000 records from the ground-truth data. These sample sizes reflected common practices in medical research, where data are often analyzed in cohorts stratified by demographic or clinical characteristics. By evaluating imputation methods across different sample sizes, we aimed to assess their robustness and scalability in handling missing data.

\subsubsection{Variable Selection and Data Characteristics}
In our experiments, we focused on a representative set of variables commonly found in EHR, including both numerical and categorical types. Specifically, we selected the following variables from the CDC dataset: ``AgeCategory,'' ``GeneralHealth,'' ``HadDiabetes,'' ``BMI,'' ``SmokerStatus,'' and ``SleepHours,'' as these variables captured key aspects of an individual's health profile, and are frequently analyzed in clinical research.

\begin{figure}[htbp]
    \centering
    \includegraphics[width=\linewidth]{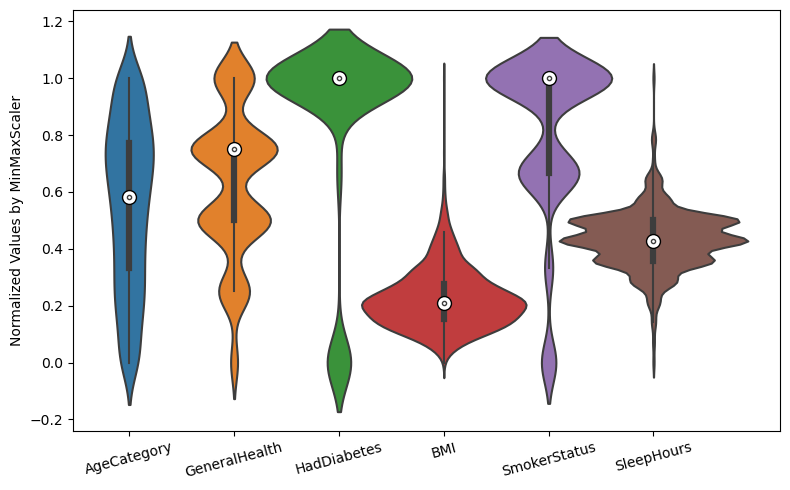}
    \caption{Distribution of the selected variables in the experimental dataset. Data is normalized within the respective ranges of variables.}
    \label{fig:data_distribution}
\end{figure}

Figure \ref{fig:data_distribution} presents the selected variable distribution in the experimental dataset, illustrating the characteristic features of numerical and categorical variables in EHR data. Numerical variables, including ``BMI'’ and ``SleepHours,'' were utilized in their original form, while categorical variables, such as ``AgeCategory'' and ``SmokerStatus,'' were encoded using ordinal labels to facilitate analysis.

\subsubsection{Experimental Methodology}
To evaluate the performance of imputation methods, we employed a controlled missingness introduction process. We treated the complete experimental datasets as the ground truth and artificially set missing values by randomly removing a fixed percentage (30\%) of data points. The missingness rate aligned with commonly observed levels of incompleteness in real-world EHR datasets.

Further, we applied SESA and various baseline imputation methods to the incomplete datasets, comparing the imputed values against the ground truth using a comprehensive set of evaluation metrics. The methodology allowed us to assess the accuracy, robustness, and efficiency of imputation methods in recovering the true underlying data distribution.

\subsubsection{Evaluation Metrics}
TTo comprehensively assess the performance of imputation methods, we employed a diverse set of evaluation metrics that captured different aspects of imputation quality, including root mean square error (RMSE), mean absolute percentage error (MAPE), coefficient of determination ($R^2$), and Wasserstein distance and Wilcoxon signed-rank test.

The evaluation metrics collectively provided a comprehensive assessment of imputation methods, capturing their accuracy, distributional matching, and statistical significance. By employing a diverse set of metrics, we aimed to provide a robust and reliable evaluation of SESA and the baseline methods.

\subsubsection{Baseline Imputation Methods}
To benchmark the SESA performance, we compared SESA against a range of widely used imputation methods, including both traditional statistical approaches and ML-based techniques. The baseline imputation methods considered in our experiments included Mean, Median, K-Neighbors, Random Forest, Bayesian Ridge Regression, Gradient Boosting, Epsilon-Support Vector Imputation (SVR), and Multilayer Perceptron (MLP).

These baseline methods represented a diverse range of imputation techniques, encompassing both traditional statistical approaches and advanced ML algorithms. By comparing SESA against the aforementioned methods, we aimed to demonstrate its effectiveness and superiority while handling missing data in EHR.

\subsubsection{Experimental Environment}

The experimental setup employed a Python environment (version 3.8 or higher) for data preprocessing utilizing Pandas. Baseline methods were implemented using the scikit-learn library. Causal analysis was conducted via the CausalNex library, operating in a CPU environment. The SESA algorithm was developed using the PyTorch framework and executed on NVIDIA RTX A6000 GPU.

\subsection{Experimental Results}
In this section, we present the experimental results, evaluating the performance of SESA and the baseline imputation methods across different sample sizes and missingness scenarios.

\subsubsection{Imputation Performance Across Sample Sizes}
Table \ref{tab:results_sample_size} lists the imputation performance results of SESA and the baseline methods across different sample sizes (300, 1000, and 3000) under a fixed missingness rate of 30\%. The evaluation metrics, including RMSE, MAPE, $R^2$, and Wasserstein distance, provided a comprehensive assessment of the imputation quality.

The results demonstrated the superior performance of SESA compared to the baseline methods, particularly at smaller sample sizes. At a sample size of 300, SESA achieved the lowest RMSE and MAPE values and the highest $R^2$, indicating its ability to accurately impute missing values even with limited data. As the sample size increased to 1000 and 3000, SESA maintained its competitive performance, consistently ranking among the top methods across all evaluation metrics.

Notably, the mean and median imputation methods performed reasonably well in certain scenarios despite their simplicity, particularly at larger sample sizes. However, their performance was less consistent across different evaluation metrics, and they often yielded higher Wasserstein distances,  suggesting a poorer match to the true data distribution .

ML-based imputation methods, such as KNN, Random Forest, Bayesian Ridge, Gradient Boosting, SVR, and MLP, generally exhibited more robust and consistent performance across sample sizes. Among these methods, Bayesian Ridge, SVR, and MLP tended to outperform others in terms of RMSE, MAPE, and $R^2$ at larger sample sizes. KNN, Random Forest, and Gradient Boosting consistently achieved lower Wasserstein distances, indicating a better alignment with the true data distribution.

\begin{table*}
     \caption{Imputation performance across different sample sizes (30\% missingness rate)}
    \label{tab:results_sample_size}
  \centering
  \begin{tabular}{lr>{\raggedleft\arraybackslash}p{1.5cm}>{\raggedright\arraybackslash}p{0.5cm}>{\raggedleft\arraybackslash}p{1.5cm}>{\raggedright\arraybackslash}p{0.5cm}>{\raggedleft\arraybackslash}p{1cm}>{\raggedright\arraybackslash}p{0.5cm}>{\raggedleft\arraybackslash}p{2.1cm}>{\raggedright\arraybackslash}p{0.5cm}}
    \toprule
   \textbf{ Imputation methods}& \textbf{Sample size}&\textbf{ RMSE }&  &\textbf{ MAPE\%} &  &\textbf{ $R^2$} && \textbf{Wasserstein-Dist} & \\
    
    \midrule
    Mean Imputation&300 & 1.2487& & 11.2723 && 0.6945 && 0.5723 &\\
    &1000 & 1.3007&$\bullet$& 10.8440&$\bullet$&0.6639 &$\bullet$&0.5337 &\\
    &3000 &1.3485&& 10.4587&& 0.6861&$\bullet$& 0.5536 &\\

    \midrule
    Median Imputation&300 & 1.2474 &$\bullet$& 12.2662 && 0.6762 && 0.5101& \\
    &1000 &  1.3217&$\bullet$&11.4416& &0.6487 &&0.5159& \\
    &3000 & 1.3686&& 11.0255&& 0.6712&& 0.5348 &\\

    \midrule
    K-Neighbors Imputation&300  & 1.3925 && 11.0388 &$\bullet$& 0.6621 && 0.2581&$\bullet$ \\
    &1000 & 1.4498&&11.2692 && 0.6011 &&0.2341 &$\bullet$\\
    &3000 &1.4586&& 10.5035&& 0.6414&& 0.2068 &$\bullet$\\

    \midrule
    Random Forest Imputation&300 & 1.4311 && 10.9684 &$\bullet$& 0.6260 && 0.2242 &$\bullet$\\
    &1000 &1.4628&&10.9906 &&0.5956 &&0.2321& $\bullet$\\
    &3000 &1.4889&& 10.3550&& 0.6330&& 0.2272&$\bullet$ \\

    \midrule
    Bayesian Ridge Imputation&300 & 1.3101 && 11.4200 &&  0.6809 & $\bullet$&   0.5087 &\\
             &1000 & 1.3299 &&  11.1335 &&  0.6504  &&    0.4643& \\
        &3000 & 1.3188&$\bullet$& 9.8909&$\bullet$& 0.7117&$\bullet$& 0.3766 &\\    

    \midrule
    Gradient Boosting Imputation&300  &1.4214 && 11.1105 && 0.6364 &&  0.2344& $\bullet$\\
    &1000 & 1.4013 &&  10.9240 &$\bullet$&  0.6177 &&   0.2718&$\bullet$ \\
    &3000 & 1.4234 && 10.0576 &$\bullet$&  0.6570 &&    0.2503&$\bullet$ \\
    
        \midrule
    Epsilon-Support Vector Imputation&300 &1.2406 &$\bullet$& 11.9456 &&  0.6871  &&  0.4356 &\\
                                          &1000 & 1.3113 &&  11.1278& & 0.6576  &$\bullet$&    0.4006& \\
                                           &3000 & 1.3304 &$\bullet$&  10.4246 &&  0.6916 &$\bullet$&   0.3969& \\

        \midrule
    Multilayer Perceptron Imputation&300 & 1.6714 && 13.5015 &&  0.5215&&    0.2606 &\\
                                       &1000 & 1.3920 &&  10.9748 &&  0.6453  &&    0.2995& \\
                                       &3000 &1.3660 &$\bullet$& 9.9485 &$\bullet$&  0.6890   &$\bullet$&    0.3144&\\

    \bottomrule
    SESA (BMI model) &300 & 1.2120 &$\bullet$& 10.9471 &$\bullet$& 0.7172 &$\bullet$& 0.3831& \\
    &1000 & 1.2822&$\bullet$&10.4313 &$\bullet$&0.6828 &$\bullet$&0.3671 &\\
             &3000 &1.4085 &&  10.6284 &&  0.6663 & &     0.4415 &\\
    \midrule
    \midrule
    SESA (GeneralHealth model) &300 & 1.2543 && 12.3369 && 0.6521 && 0.4309& \\
    (for Section~\ref{sec:disc} only)&1000 & 1.3345 && 10.6692&&  0.6822 &&     0.4695 &\\
             &3000 &1.3682 &&  10.9475 &&  0.6611 & &   0.4831 &\\

    \bottomrule
 
    \multicolumn{10}{l}{\footnotesize Note: (1) The metrics of RMSE, MAPE\%, $R^2$, and Wasserstein Distance represent the mean values aggregated across all assessed variables, providing}\\
     \multicolumn{10}{l}{\footnotesize  a view of the imputation performance. (2)  The symbol ($\bullet$) denotes the methods that ranked within the top three except SESA (GeneralHealth model)}  \\
        \multicolumn{10}{l}{\footnotesize for performance, considering the specified sample size, thereby highlighting the most effective imputation strategies in the given context.}\\
  \end{tabular}

  \label{tab:Exp-1.1}
\end{table*}

\subsubsection{Imputation Performance by Variable}
To gain a more detailed understanding of the imputation performance, we examined the results for each variable separately. Table \ref{tab:Exp-1.2} presents the Wilcoxon signed-rank test results, comparing the imputed values against the true values for each variable at a sample size of 1000 and a missingness rate of 30\%.

The Wilcoxon signed-rank test assesses the statistical significance of the differences between the imputed and true values, with a significance level of 0.05. A high p-value (close to 1.0) indicates insufficient evidence to reject the null hypothesis, suggesting no significant differences between the imputed and true values. Effect size and confidence intervals provide additional insights into the magnitude and precision of the imputation results.

The results highlighted the strong performance of SESA and Bayesian Ridge across different variable types. For variables such as ``GeneralHealth,'' ``HadDiabetes,'' and ``BMI,'' SESA and Bayesian Ridge consistently achieved high p-values, indicating a close match between the imputed and true values. The small differences in the Wilcoxon rank statistics and effect sizes among the methods suggested their robustness in handling both numerical and categorical variables.

For the ``SmokerStatus'' variable, Gradient Boosting emerged as the top-performing method, followed by SESA and Random Forest. Larger differences in the Wilcoxon rank statistics for other methods were attributable to the complex relationships between ``SmokerStatus'' and other variables, which Gradient Boosting captured effectively.

Overall, the variable-level analysis reinforced the effectiveness of SESA in imputing missing values across different data types and distributions. The favorable Wilcoxon test results and the consistency of SESA's performance with other top-ranking methods underscore its reliability and versatility for handling missing data in EHR.

\begin{table*}
\caption{Imputation performance by variable (sample size: 1000, missingness rate: 30\%)}
  \centering
  \begin{tabular}{ll>{\raggedleft\arraybackslash}p{3cm}>{\raggedleft\arraybackslash}p{1.5cm}>{\raggedright\arraybackslash}p{1cm}>{\raggedleft\arraybackslash}p{1.5cm}>{\raggedleft\arraybackslash}p{1cm}>{\raggedright\arraybackslash}p{1cm}}
    \toprule
   \textbf{Variables}& \textbf{Imputation methods}& \textbf{Wilcoxon Rank Statistic} & \textbf{p-value} && \textbf{Effect Size} &   \multicolumn{2}{c}{\textbf{95\% CI}} \\
    & &  &(0.05) && & lower & upper \\    
    \midrule
    AgeCategory& Mean& 20992.5 & 0.7828 &$\bullet$& 663.8411& 8.0 & 9.0 \\        
                        & Median& 20992.5 & 0.7828 &$\bullet$& 663.8411& 8.0 & 9.0  \\        
                        & K-Neighbors &  17117.5  & 0.0068 &&  541.3029 &8.0 & 9.0 \\
                        & Random Forest & 16607.0 & 0.0144 &&   525.1595 & 8.0 & 9.0 \\
                        & Bayesian Ridge &  17104.5 & 0.1063 &&  540.8918 & 8.0 & 9.0 \\                    
                        & Gradient Boosting & 19090.5 & 0.2231 &&   603.6946 & 8.0 & 9.0 \\                         
                        & Epsilon-Support Vector  & 19910.0  & 0.6615 &$\bullet$&  629.6095  & 8.0 & 9.0 \\    
                        & Multilayer Perceptron  & 19613.0 & 0.2969 &&  620.2175  & 8.0 & 9.0 \\    
                        & SESA (BMI model) & 20067.0  & 0.1936 &&  634.5743 & 8.0 & 9.0 \\    
                        & SESA (GeneralHealth model) &  12531.5  & 0.0281 &&  396.2808 & 8.0 & 8.0 \\    
                                                                                
    \midrule
    GeneralHealth& Mean& 4631.5  & $<0.0001$  & & 146.4609& 4.0 & 4.0 \\        
                        & Median&  4161.0  & $<0.0001$  & & 131.5824 & 4.0 & 4.0 \\        
                        & K-Neighbors & 10283.5 & 0.8308 &$\bullet$&   325.1928 & 4.0 & 4.0 \\
                        & Random Forest & 7430.5 & 0.0130 &&  234.9730 & 4.0 & 4.0 \\
                        & Bayesian Ridge &  9020.0  & 0.6376 &$\bullet$&   285.2374    & 4.0 & 4.0 \\                    
                        & Gradient Boosting &8600.0  & 0.2466 &&   271.9560 & 4.0 & 4.0 \\                         
                        & Epsilon-Support Vector  &  6935.0 & 0.0487 &&  219.3040 & 4.0 & 4.0 \\    
                        & Multilayer Perceptron  & 7728.0 &  0.0997 &&   244.3808 & 4.0 & 4.0 \\    
                        & SESA  (BMI model)& 6355.0  & 0.3877 &$\bullet$&   200.9627 & 4.0 & 4.0\\    
                       & SESA (GeneralHealth model) &  7963.0  & 0.5013 &&   251.8122 & 4.0 & 4.0 \\    
                        
    \midrule
    HadDiabetes& Mean&  0.0  & $<0.0001$  &&   0.0000&  4.0 & 4.0 \\        
                        & Median&  0.0 &  $<0.0001$   &&    0.0000 &  4.0 & 4.0 \\        
                        & K-Neighbors &  1972.5 & 0.1390 &$\bullet$&   62.3759 &  4.0 & 4.0 \\
                        & Random Forest & 3726.0 & 0.0269 && 117.8265 &  4.0 & 4.0 \\
                        & Bayesian Ridge & 4918.0 & 0.5265 &$\bullet$& 155.5208  &  4.0 & 4.0 \\                    
                        & Gradient Boosting & 3798.0 & 0.0037 &&  120.1033  &  4.0 & 4.0 \\                         
                        & Epsilon-Support Vector  &  0.0 &  $<0.0001$ & &   0.0000 & 4.0 & 4.0\\    
                        & Multilayer Perceptron  & 5738.5 & 0.0588 & & 181.4673  &  4.0 & 4.0 \\    
                        & SESA  (BMI model)&8163.0  & 0.2064 &$\bullet$&   258.1367   & 4.0 & 4.0 \\    
                        & SESA (GeneralHealth model) &  4085.5  & 0.1495 &&   129.1949 & 4.0 & 4.0 \\     
    \midrule
    BMI& Mean& 20493.0 & 0.6016 &&   648.0456& 27.11 &  27.98 \\        
                        & Median& 17371.5 & 0.0091& &   549.3351 &  27.10   &  27.97  \\        
                        & K-Neighbors & 20317.0 & 0.5192 &&  642.4800  & 27.12   &   27.94  \\
                        & Random Forest & 20388.0 & 0.5518 &&  644.7252  & 27.10 &  27.98  \\
                        & Bayesian Ridge & 20520.0  & 0.6148 &$\bullet$&  648.8994    &27.11   &  27.98   \\                    
                        & Gradient Boosting & 19474.0 &  0.2182 &&  615.8220 & 27.12    &    27.98   \\                         
                        & Epsilon-Support Vector  & 17672.0 & 0.0129 &&   558.8378 & 27.11     &    27.98  \\    
                        & Multilayer Perceptron  & 20918.0 & 0.8210 &$\bullet$&  661.4852 & 27.11    &    27.98  \\    
                        & SESA  (BMI model)&21746.0  & 0.7221 &$\bullet$ &  687.6689 & 27.10  &   27.98 \\    
                        & SESA  (GeneralHealth model)&20335.0  & 0.0215 & &  643.0492   & 27.21 &   28.13 \\    

    \midrule
    SmokerStatus& Mean& 7440.0 & $<0.0001$   && 235.2735  &4.0  &  4.0  \\        
                        & Median&  0.0 &$<0.0001$ & &   0.0000 & 4.0  &  4.0 \\        
                        & K-Neighbors &  5166.0 &  0.2096 & & 163.3633  & 4.0  &  4.0 \\
                        & Random Forest & 7539.5  & 0.4336 &$\bullet$&   238.4199   &  4.0  &  4.0  \\
                        & Bayesian Ridge & 8078.5  & 0.2464  && 255.4646  &4.0  &  4.0 \\                    
                        & Gradient Boosting & 8084.0 & 0.9258 &$\bullet$&  255.6385  & 4.0  &  4.0  \\                         
                        & Epsilon-Support Vector  &  0.0 &  $<0.0001$  & &   0.0000  & 4.0  &  4.0   \\    
                        & Multilayer Perceptron  & 7758.0 &  0.3907 &&  245.3295 & 4.0  &  4.0 \\    
                        & SESA (BMI model) &  6921.0  &0.3946 &$\bullet$&   218.8612 & 4.0  &  4.0   \\   
                        & SESA (GeneralHealth model) &  7024.5  &0.5907 &&   222.1342 & 4.0  &  4.0   \\                           

    \midrule
    SleepHours& Mean& 8894.0 & 0.8044 &$\bullet$&   281.2530&  7.0  &  7.0  \\        
                        & Median& 8894.0  & 0.8044 &$\bullet$&   281.2530  & 7.0  &  7.0 \\        
                        & K-Neighbors &  9906.0 & 0.5033 &&   313.2552  & 7.0  &  7.0 \\
                        & Random Forest & 9471.0 & 0.3961 &&  299.4993  & 7.0  &  7.0 \\
                        & Bayesian Ridge &8356.0 & 0.6269 &$\bullet$&   264.2399 & 7.0  &  7.0 \\                    
                        & Gradient Boosting & 8920.5  & 0.5580 &&   282.0910   & 7.0  &  7.0 \\                         
                        & Epsilon-Support Vector  & 9223.0  & 0.9553 &$\bullet$&  291.6569  & 7.0  &  7.0 \\    
                        & Multilayer Perceptron  & 8614.5  & 0.3170&  &  272.4144 & 7.0  &  7.0 \\    
                        & SESA (BMI model) & 6921.0  & 0.3946& &   218.8612      & 7.0  &  7.0 \\    
                        & SESA (GeneralHealth model) & 9251.0  & 0.0404& &   292.5423      & 7.0  &  7.0 \\    
                        
    \bottomrule
 
    \multicolumn{1}{r}{\footnotesize Note:} & \multicolumn{7}{l}{\footnotesize The symbol ($\bullet$) identifies the imputation methods ranked within the top three except SESA (GeneralHealth model).}  \\
    \multicolumn{1}{r}{\footnotesize } & \multicolumn{7}{l}{\footnotesize   This ranking implies insufficient evidence to reject the null hypothesis, suggesting no statistically significant differences}  \\
        \multicolumn{1}{r}{\footnotesize } & \multicolumn{7}{l}{\footnotesize  between the imputation results and the original data.}  
  \end{tabular}

  \label{tab:Exp-1.2}
\end{table*}

\subsection{Improving SEM Initialization Based on Causal Discovery Directed Acyclic Graph (DAG)}

\begin{figure}[h]\centering 
\includegraphics[width=1.0\linewidth]{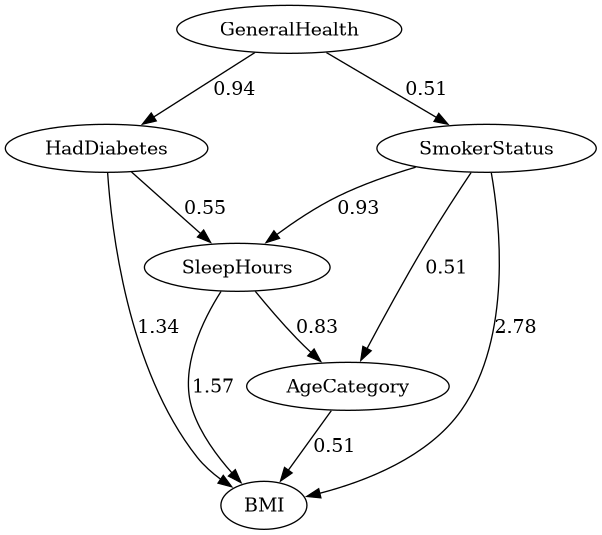}
\caption{Causal discovery analysis results of for the selected variables group by NOTEARS algorithm. In this context, nodes represent the variables included in the dataset, while directed edges indicate potential causal directions. The weight of each directed edge reflects the strength of the causal effect.}
\label{fig:NOTEARS_0.5}
\end{figure}

Following the aforementioned discussion, when the optimal initialization model could not be determined through SEM fitting and analysis, we considered employing causal discovery methods to seek better solutions. Using the NOTEARS algorithm for directed acyclic graph (DAG) analysis of selected variables, we aimed to gain a more rational SEM initialization input from the causal structure. The NOTEARS algorithm\cite{NOTEARS2018} is an optimization-based method designed to learn DAGs from data, ensuring an acrylic learned graph structure.

With the same dataset and experimental setup, the results from the NOTEARS algorithm revealed potential causal relationships between variables, uncovering possible hierarchical structures and providing estimates of causal direction and strength. Particularly, the DAG analysis reinforced the rationale behind the BMI model as an initialization input for SEM.

Tables\ref{tab:Exp-1.1} and \ref{tab:Exp-1.2} present the results of SESA (BMI model) and SESA (GeneralHealth model). The experimental findings indicated that the former, which aligned more closely with the DAG analysis, slightly outperformed the latter across almost all evaluation metrics. Two interesting observations emerged from the analysis: Firstly, the latter model exhibited better imputation performance for GeneralHealth than the former during the Wilcoxon rank test, indirectly validating the impact of initial SEM input on the outcome of SESA. Secondly, at a large data scale (3000 entities), the discrepancy between the two different initial SEM models in terms of RMSE, MAPE, $R^2$, and Wasserstein distance was significantly reduced compared to smaller and medium data scales. The phenomenon suggests the enhanced role of the self-attention mechanism; as the dataset size increases, self-attention is capable of more substantially mitigating the disparities caused by different SEM model initializations.

On comparing the GeneralHealth model as an initialization, results indicated that the BMI model was closer to the causal relationships parsed by the NOTEARS algorithm, thus providing better SEM fitting and FIML estimation outcomes. Additionally, the self-attention mechanism exhibited positive optimization effects on the FIML estimation results under both SEM configurations, making certain corrections even under the less reasonable SEM setup.

IIn summary, the statistical-based FIML estimation and the neural network-based self-attention mechanism demonstrate complementary advantages. FIML estimation proves robustness in small data environments, providing a solid foundation for self-attention, which leverages the learning strengths of neural networks in medium-sized data environments to uncover potential deep relationships in data, thereby optimizing imputation results.

However, improvements in SEM models are not without constraints due to data heterogeneity and locality, which are essential to acknowledge. Data heterogeneity refers to the diversity and variability among individuals or groups within a dataset, leading to a variety of characteristics and behaviors. Locality implies that data analysis or model construction can only consider specific variables or subsets, potentially overlooking broader contexts or other relevant variables. Inductive analyses of SEM and DAG are based on population-scale generalizations, unable to cover individual differences and all potential variable relationships comprehensively. Therefore, selecting appropriate variable groups for data imputation is crucial, based on prior knowledge and research objectives, to aptly reflect the problem structure while minimizing bias introduction.

\section{Discussion}
\label{sec:disc}

The experimental results and analyses presented in this study provide compelling evidence for the effectiveness and superiority of the SESA method while handling missing data in EHR. By synergistically integrating SEM and self-attention mechanisms, SESA offers a principled and adaptive approach to imputation that captures the complex relationships and dependencies among variables in EHR data.

\subsection{Insights from SEM Fitting and Analysis}
The fitting and analysis of SEM models play a crucial role in the SESA imputation method \cite{Allison2003}. The complexity of variable relationships in EHR data poses challenges for conventional correlation analysis and SEM model testing in uncovering deeper connections within the data. In this study, we propose two SEM models for analysis: the BMI and GeneralHealth models. These models are fitted and analyzed using the Python semopy library \cite{semopy2020} to assess the relationships between the dependent variables and other variables and to evaluate the model fit.

The SEM analytical results, presented in Table \ref{tab:SEM_1}, reveal significant associations between the dependent variables (BMI and GeneralHealth) and various independent variables, such as AgeCategory, SleepHours, HadDiabetes, and SmokerStatus. The estimates, standard errors, and p-values indicate the direction, magnitude, and statistical significance of these relationships. The excellent fit indices, including the comparative fit index (CFI) and RMSE of approximation (RMSEA), suggest that both models are good fits for the data.

However, the SEM analysis alone does not provide a definitive answer as to which model is better for initializing SESA imputation. Thus, the challenge highlights the need for additional techniques to guide the selection of the most appropriate SEM model for initialization.

\begin{table*}
\centering
\caption{Parameter estimates for SEM (BMI model) and SEM (GeneralHealth model)}
  \begin{tabular}{lc>{\raggedright\arraybackslash}p{3.0cm}>{\raggedleft\arraybackslash}p{2.0cm}>{\raggedleft\arraybackslash}p{2.0cm}>{\raggedleft\arraybackslash}p{2.0cm}}
   \toprule
\textbf{Dependent Variable} && \textbf{Independent Variable} & \textbf{Estimate} & \textbf{Std. Err} & \textbf{p-value} \\
\hline
BMI &$\sim$& GeneralHealth & -1.3283 & 0.1150 & $<0.0001$ \\
BMI &$\sim$& AgeCategory & -0.1060 & 0.0336 & 0.0016 \\
BMI &$\sim$& SleepHours & -0.2146 & 0.0769 & 0.0052 \\
BMI &$\sim$& HadDiabetes & -1.1532 & 0.1128 & $<0.0001$ \\
BMI &$\sim$& SmokerStatus & 0.2503 & 0.1266 & 0.0480 \\
BMI &$\sim$ $\sim$& BMI & 38.4322 & 0.9923 & $<0.0001$ \\
\hline
GeneralHealth &$\sim$& BMI & -0.0320 & 0.0028 & $<0.0001$ \\
GeneralHealth &$\sim$& AgeCategory & -0.0238 & 0.0052 & $<0.0001$ \\
GeneralHealth &$\sim$& SleepHours & 0.0352 & 0.0119 & 0.0032 \\
GeneralHealth &$\sim$& HadDiabetes & 0.1784 & 0.0175 & $<0.0001$ \\
GeneralHealth &$\sim$& SmokerStatus & 0.2001 & 0.0193 & $<0.0001$ \\
GeneralHealth &$\sim$ $\sim$& GeneralHealth & 0.9280 & 0.0240 & $<0.0001$ \\
\hline
\multicolumn{2}{l}{Results for BMI model: } &\multicolumn{4}{l}{Optimization result: 7.5608e-08;~~CFI: 1.0188;~~RMSEA: 0} \\
\multicolumn{2}{l}{Results for GeneralHealth model:} &\multicolumn{4}{l}{Optimization result: 1.1269e-07;~~CFI: 1.0188;~~RMSEA: 0} \\
\bottomrule
\end{tabular}
\label{tab:SEM_1}
\end{table*}

\subsection{Enhancing SEM Initialization through Causal Discovery}
To address the challenge of selecting an optimal SEM model for initialization, we propose the integration of causal discovery methods, specifically the NOTEARS algorithm \cite{NOTEARS2018}, into the SESA framework. By employing DAG analysis on the selected variables, we aim to gain insights into the causal structure underlying the data, informing the choice of SEM model for initialization.

The results of the NOTEARS algorithm, visualized in Fig. \ref{fig:NOTEARS_0.5}, reveal potential causal relationships among the variables, uncovering hierarchical structures and providing estimates of causal direction and strength. Notably, the DAG analysis supports the rationale behind the BMI model as a suitable initialization input for SEM.

The experimental findings as presented in Tables \ref{tab:Exp-1.1} and \ref{tab:Exp-1.2} demonstrate that the SESA method initialized with the BMI model, which aligns more closely with DAG analysis, consistently outperforms the SESA method initialized with the GeneralHealth model across various evaluation metrics. Our observation underscores the importance of incorporating causal discovery techniques to guide the selection of the most appropriate SEM model for initialization.

Interestingly, the Wilcoxon rank test results reveal that the GeneralHealth model shows better imputation performance for the GeneralHealth variable compared to the BMI model. Thus, the finding suggests that the choice of SEM model for initialization can have a variable-specific impact on the imputation outcomes. Furthermore, as the dataset size increases, the discrepancy between the two SEM initialization models diminishes, indicating the enhanced role of the self-attention mechanism in mitigating the differences caused by different SEM initializations.

\section{Conclusion}
\label{sec:Conclusion}

SESA is an effective method for addressing the critical challenge of missing data in EHR. By synergistically integrating SEM and self-attention mechanisms, SESA offers a principled, adaptive, and robust approach to imputation that outperforms traditional methods.

Grounded in statistical learning and information theories, SESA leverages SEM to capture complex relationships among variables in EHR data, while the self-attention mechanism enables adaptive learning of intricate patterns. Comprehensive experiments across real-world EHR datasets and various missingness scenarios establish the efficacy and superiority of SESA, consistently outperforming state-of-the-art imputation methods in terms of accuracy, robustness, and versatility.

From an information--theoretic perspective, self-attention can be viewed as a mechanism for maximizing the mutual information between input features and imputed values. By learning attention weights that prioritize informative features, SESA effectively reduces the uncertainty associated with missing values and improves imputation quality.

In conclusion, SESA represents a substantial advancement in missing data imputation, particularly within the context of EHR. The innovative integration of SEM and self-attention mechanisms, coupled with the strong theoretical foundations and empirical effectiveness, positions SESA as a valuable tool for researchers and practitioners.

\section{Limitation and Future Work}
\label{sec:Limitation}
While SESA demonstrates advancements while handling missing data in EHR, acknowledging its limitations and identifying potential areas for future improvements are essential. One current limitation is its reliance on the linear assumptions of SEM, which may not fully capture the complex, non-linear relationships often present in healthcare data. Future research can explore the integration of non-linear SEM models or the incorporation of non-linear dynamics within the SESA framework to better adapt to the intricacies and heterogeneity of real-world EHR datasets.

Another challenge includes the risk of overfitting, particularly when dealing with high-dimensional EHR data. Future versions of SESA can incorporate robust regularization techniques and advanced validation strategies to mitigate the issue and maintain its effectiveness across a wider range of datasets and missingness scenarios.

With increasing healthcare data volume and complexity, advanced imputation techniques including SESA become increasingly crucial. Future research directions include extending SESA to handle diverse data types, incorporating temporal and causal modeling, and integrating it with other healthcare applications. Through continued innovation and interdisciplinary collaboration, we can push the boundaries of missing data imputation and harness the power of EHR to improve patient outcomes and advance healthcare research.

\section*{Acknowledgments}
The work was supported in part by the 2023-2024 Waseda University Advanced Research Center Project for Regional Cooperation Support, 2020-2025 JSPS A3 Foresight Program (Grant No. JPJSA3F20200001), 2022–2024 Japan National Initiative Promotion Grant for Digital Rural City, 2022-2024 Masaru Ibuka Foundation Research Project on Oriental Medicine, 2023 and 2024 Waseda University Grants for Special Research Projects (Nos. 2023C-216 and 2024C-223), and 2023-2024 Japan Association for the Advancement of Medical Equipment (JAAME) Grant.

\bibliographystyle{IEEEtran}
\bibliography{references}

\end{document}